\documentclass[letterpaper, 10 pt, conference]{IEEEtran}
\IEEEoverridecommandlockouts
\usepackage{cite}
\usepackage{amsmath,amssymb,amsfonts}
\usepackage{algorithmic}
\usepackage{graphicx}
\usepackage{textcomp}
\usepackage{xcolor}
\usepackage[lofdepth,lotdepth]{subfig}
\usepackage{hyperref}
\usepackage{caption}
\newcommand{\specialcell}[2][c]{%
  \begin{tabular}[#1]{@{}c@{}}#2\end{tabular}}
  
\def\BibTeX{{\rm B\kern-.05em{\sc i\kern-.025em b}\kern-.08em
    T\kern-.1667em\lower.7ex\hbox{E}\kern-.125emX}}
\begin{document}


\title{\raisebox{5.9mm}{\strut} COLA: COarse LAbel pre-training for 3D semantic segmentation of sparse LiDAR datasets 
}

\author{\IEEEauthorblockN{Jules Sanchez$^1$}
\and 
\IEEEauthorblockN{Jean-Emmanuel Deschaud$^1$}
\and 
\IEEEauthorblockN{François Goulette$^{2,1}$}
\and
\IEEEauthorblockA{$^1$\textit{Centre for Robotics, Mines Paris - PSL, PSL University,} \\
75006 Paris, France \\
firstname.surname@minesparis.psl.eu}
\and
\IEEEauthorblockA{$^2$\textit{U2IS, ENSTA Paris, Institut Polytechnique de Paris,} \\
91120 Palaiseau, France \\
firstname.surname@ensta-paris.fr}
}

\maketitle

\begin{abstract}
Transfer learning is a proven technique in 2D computer vision to leverage the large amount of data available and achieve high performance with datasets limited in size due to the cost of acquisition or annotation. 
In 3D, annotation is known to be a costly task; nevertheless, pre-training methods have only recently been investigated. Due to this cost, unsupervised pre-training has been heavily favored. 

In this work, we tackle the case of real-time 3D semantic segmentation of sparse autonomous driving LiDAR scans. Such datasets have been increasingly released, but each has a unique label set. We propose here an intermediate-level label set called coarse labels, which can easily be used on any existing and future autonomous driving datasets, thus allowing all the data available to be leveraged at once without any additional manual labeling.

This way, we have access to a larger dataset, alongside a simple task of semantic segmentation. With it, we introduce a new pre-training task: coarse label pre-training, also called COLA. 

We thoroughly analyze the impact of COLA on various datasets and architectures and show that it yields a noticeable performance improvement, especially when only a small dataset is available for the finetuning task. 
\end{abstract}

\begin{IEEEkeywords}
LiDAR, semantic segmentation, pre-training
\end{IEEEkeywords}

\section{Introduction}

LiDAR-based deep learning has been gaining a lot of traction in recent years, especially for autonomous driving. Thus, an increasing amount of data has been released for scene understanding, such as KITTI \cite{Geiger2012CVPR} and Argoverse 2 \cite{Argoverse2} for detection or SemanticKITTI \cite{behley2019iccv} and nuScenes \cite{nuscenes2019} for semantic segmentation.

So far, 3D semantic segmentation methods have focused on architecture design and tuning rather than data-driven approaches to improve performance. Some works have started to emerge on \textit{domain adaptation} and \textit{transfer learning} to ensure good performances when the amount of target annotations is limited or nonexistent. 

Transfer learning is the ability to reuse previously learned information in a deep learning model.  We can decompose transfer learning into two parts: first, a pre-training task for which there is usually a large amount of data available, and then a finetuning task, which is the task we want our model to achieve in the end. The pre-training task helps the network to learn useful representations that should speed up the training and improve the results of the finetuning task.

Domain adaptation is a subset of transfer learning, aiming specifically to transfer knowledge when the pre-training and finetuning tasks are similar. However, it is applied in a specific scope where the pre-training data and the finetuning data are available simultaneously. 

\begin{figure}[t]
        \centering
        \includegraphics[width=\linewidth]{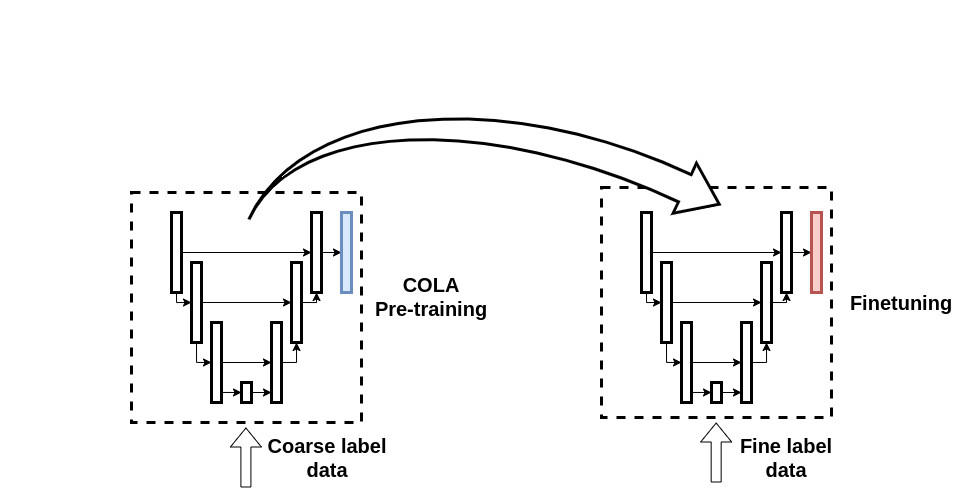}
\\
        \subfloat[SemanticKITTI coarse labels. \\ 8 different labels]{
            \includegraphics[width=.45\linewidth]{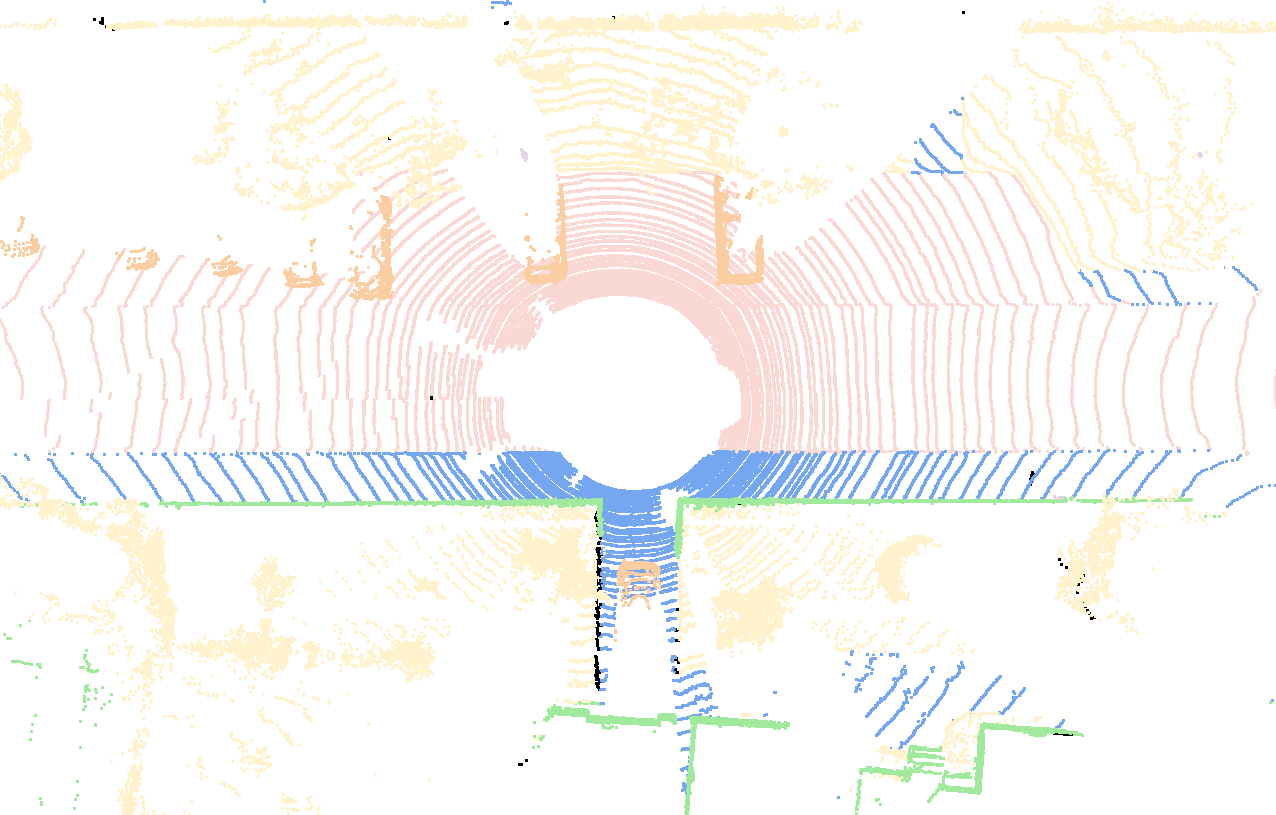}
        }
        \subfloat[SemanticPOSS fine labels. \\ 13 different labels]{
            \includegraphics[width=.45\linewidth]{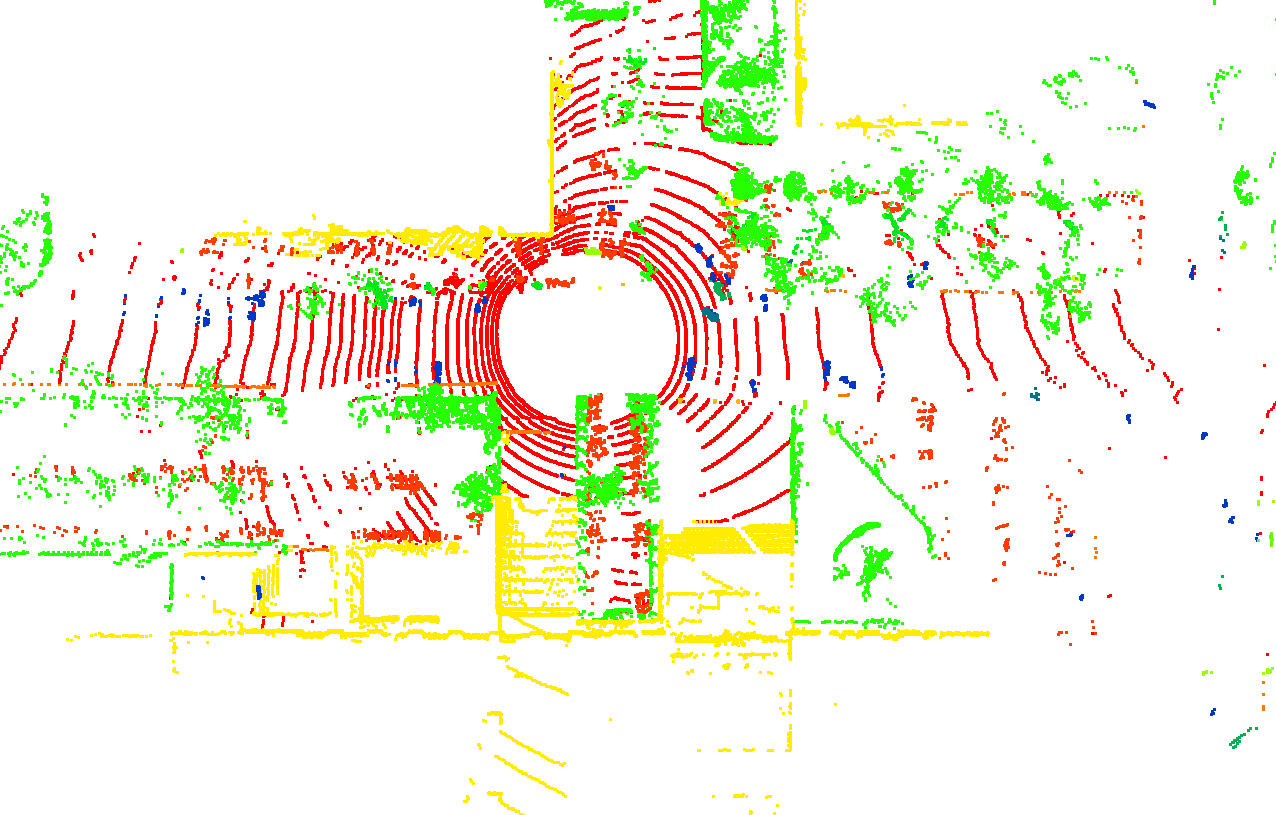}
        }
        \caption{COLA pipeline.}
        \label{fig:pipe}
\end{figure}

Thus far, most of the work done in 3D pre-training has been leveraging contrastive learning (see \autoref{subsec:tf}).

We propose a new pre-training method that leverages a set of heterogeneous source datasets thanks to a simple label unification techniques for LiDAR point cloud semantic segmentation. Contrary to domain adaptation methods, access to the pre-training data is not required when finetuning our models. This way, we can provide off-the-shelf pre-trained models that can be directly finetuned on any autonomous driving semantic segmentation dataset and systematically increase performance on the finetuning task.

We introduce an intermediate-level label set called coarse labels that can be applied to any autonomous driving semantic segmentation dataset. We demonstrate that using this intermediate label set enables us to use all datasets together to pre-train neural architectures and improve performance after finetuning.

Furthermore, we highlight the limitation of existing contrastive-based methods for the task of autonomous driving semantic segmentation by looking at CSC results. For a fairer comparison, we also train our own contrastive learning-based method by following PointContrast \cite{xie2020pointcontrast} and applying it to sparse autonomous driving data. We call it SPC, for Sparse PointContrast.

The mapping from existing available datasets to the proposed label set alongside links to our pre-trained models using the largest source dataset can be found at \url{https://github.com/JulesSanchez/CoarseLabel}.

In practice, our contributions are as follows:

\begin{itemize}
    \item Studying the limitations of current contrastive based methods for autonomous driving 3D semantic segmentation.
    \item Introducing a label set specific to the task of semantic segmentation of autonomous driving (sparse LiDAR scan) generic enough to leverage every dataset at once.
    \item Thoroughly experimenting on a new pre-training that leverages this new label set. This method, called COLA (COarse LAbel pre-training) is shown in \autoref{fig:pipe}. It is quantitatively and qualitatively studied, showing a mIoU improvement of up to +6.4\% on the target dataset.
\end{itemize}

\section{Related Works}
\label{sec:sota}

\subsection{3D autonomous driving semantic segmentation}

Semantic segmentation for autonomous driving has two main challenges: achieving real-time inference and dealing with the unordered structure of the LiDAR point clouds. 

As such, several types of methods have emerged: using only a permutation invariant operation \cite{PointNet}\cite{10.5555/3295222.3295263}\cite{Hu2020RandLANetES}, redefining the convolution \cite{Tat2018}\cite{10.1007/978-3-030-01237-3_6}\cite{zhang-shellnet-iccv19}\cite{thomas2019KPConv}, projecting the point clouds in 2D with a range-based image \cite{8967762}\cite{squeezesegv3} and bird's-eye-view images \cite{Zhang_2020_CVPR} or leveraging sparse convolution-based architectures as introduced by MinkowskiNet \cite{mink}, under the name Sparse Residual U-Net (SRU-Net). 

SRU-Nets have been used as the backbone for more refined methods \cite{mink}, \cite{spvnas}, \cite{Zhou2020Cylinder3DAE}, \cite{Xu2021RPVNetAD} that are the best performing methods on the various benchmarks.

\begin{figure*}[h]
    \centering
    \includegraphics[width=0.9\textwidth]{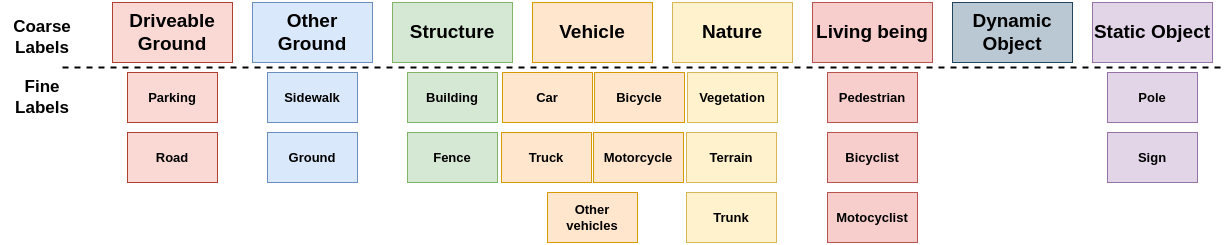}
    \caption{The proposed coarse labels and their mapping  to SemanticKITTI fine labels.}
    \label{fig:coarse_label}
\end{figure*}

\subsection{3D transfer learning}
\label{subsec:tf}

As mentioned in the introduction, 3D transfer learning has mainly relied on unsupervised contrastive pre-training tasks such as in PointContrast \cite{xie2020pointcontrast}, CSC \cite{csc2021}, DepthContrast \cite{Zhang_2021_ICCV} or
CrossPoint \cite{Afham_2022_CVPR}. PC-FractalDB \cite{Yamada_2022_CVPR} proposes an alternative to contrastive learning by learning fractal geometry as the pre-training task. Point-BERT \cite{yu2021pointbert} looks into pre-training of point cloud transformers specifically.

All those methods look into either shape understanding, indoor dense scene understanding or outdoor dense scene understanding as they rely on the large availability of such data \cite{dai2017scannet} \cite{shapenet2015} \cite{7298801} to perform their pre-training task.

Fewer methods have looked into pre-training for outdoor sparse point clouds. GCC-3D \cite{9711148} applies a two-step self-supervised method, first learning contrastive features and then leveraging the sequential nature of the data to generate pseudo instances that can be used to learn semantic clustering. ProposalContrast \cite{https://doi.org/10.48550/arxiv.2207.12654} proposes learning contrastive information at the proposal level rather than point or voxel levels. These outdoor methods are specifically designed for object detection and are not evaluated on any semantic segmentation finetuning tasks.

Finally, only SegContrast \cite{9681336} has looked into the same finetuning task as us, highlighting a gap in the study of the pre-training effect on this case.

\subsection{Leveraging heterogeneous datasets}
Using several datasets at once for supervised learning is a nontrivial task. Datasets, even for a similar task, usually have vastly different label sets. This variation can come from the type of elements in the scene or annotation choices.

To leverage as much of these data as possible, several techniques have been implemented. Some approaches have chosen to only leverage common labels across datasets \cite{Ros2016TrainingCD} \cite{9707049} or the union \cite{10.1007/978-3-030-58568-6_11}, to have one shared encoder and one decoder for each dataset \cite{10.1007/978-3-030-30645-8_28} \cite{zhou2022simple}, to leverage label relationships \cite{8659045} \cite{8500398} \cite{8578183}, finally some have chosen to unify annotations through remapping \cite{10.1007/978-3-030-33676-9_3} and re-annotation \cite{9157628}.

Category-shift multi-source domain adaptation, as defined in the proposed taxonomy of \cite{9710463}, aims to tackle a similar issue as it leverages a set of heterogeneously labeled source datasets to improve the performance of a target dataset presenting a different label set, as in Deep Cocktail Network \cite{DBLP:conf/cvpr/XuCZYL18} and mDALU \cite{9710463}.

Inspired by the first type of method, COLA focuses on leveraging the annotations of all the pre-training data without any additional human cost. Furthermore, contrary to domain adaptation methods, COLA provides off-the-shelf pre-trained models that can be finetuned without requiring access to the pre-training data.

\section{COLA}

\subsection{Shortcomings of simple contrastive approaches}
\label{sec:contrastive}

Contrastive learning has been the main approach for the pre-training of 3D models. As such, we want to investigate them for our specific finetuning task. As one of the founding and reference methods for dense point clouds, we investigate specifically CSC \cite{csc2021}.

In \autoref{tab:tab_SRU-Net_contr}, we show the results after finetuning on three major autonomous driving 3D semantic segmentation datasets when the model is pre-trained with CSC and with a LiDAR-based adaptation Sparse PointContrast (SPC). Whereas CSC is trained on dense point clouds (ScanNetV2 \cite{dai2017scannet}), SPC follows the same training methodology but with 100K LiDAR scans, leveraging mainly KITTI-360 \cite{360}. We see that CSC results in small to moderate improvement (from +$0.2\%$ to +$1.1\%$), whereas SPC pre-training systematically decreases results in the finetuning task.

We have not reported SegContrast \cite{9681336} results in \autoref{tab:tab_SRU-Net_contr}, even though it is the method working in the closest setup to ours as they are using a different model. One of the difference between SPC and SegContast is the the choice of pre-training data. While SegContrast pre-train with the same dataset as the finetuning one, here, in SPC we pre-train with datasets different from the finetuning data.

In total, the SPC pre-training took the equivalent of 50 GPU days of training. In comparison, CSC takes approximately 15 GPU days to be trained. This discrepancy stems from the difference in scan sizes regarding number of points and their range in meters. The duration of the SPC pre-training makes it unrealistic to perform it on 1M scans such as the ONCE dataset \cite{mao2021one}, which is 10 times larger than the dataset used for SPC, as it would amount to 500 GPU days.

\begin{table}[h]
\scriptsize
    \centering
    \begin{tabular}{|c|c|c|c|}
    \hline
         &  \specialcell{No pre-\\training} & \specialcell{CSC} & \specialcell{SPC}\\ \hline
        nuScenes& 67.2 & 68.3 (+1.1) & 66.3 (-0.9) \\ \hline
        SemanticKITTI& 50.5 & 50.7 (+0.2) & 48.9 (-1.6) \\ \hline
        SemanticPOSS& 55.0 & 55.5 (+0.5) & 54.4 (-0.6) \\ \hline
    \end{tabular}
    \caption{Test set mIoU for each target dataset. No pre-training, dense contrastive pre-training (CSC), and sparse contrastive pre-training (SPC) with the SRU-Net architecture.}
    \label{tab:tab_SRU-Net_contr}
\end{table}

We see that contrastive pre-training at the moment is not suited for our specific case, and scaling the method to leverage more data would require significant time and computing power. While the shortcomings of CSC, compared to results in their paper, can be explained by the variation of data topology between dense and sparse data. For SPC, we hypothesize that information at a point level is too difficult to extract properly for sparse LiDAR scans and that more meaningful structures should be used to learn insightful geometry, as it is done in SegContrast.

We believe that discarding human annotations to rely solely on unsupervised pre-training is a waste of information, and supervised methods are still relevant.

We propose a novel approach using a supervised pre-train task as opposed to the self-supervised and unsupervised-centered methods that are the typical approaches.

\subsection{Available semantic segmentation data}
Since the release of SemanticKITTI in 2019, other autonomous driving semantic segmentation LiDAR datasets have been available. We focus on five different datasets, which are the most important at the time of the writing due to the variety in size, scene types, and hardware used for acquisition. As seen in \autoref{tab:dataset}, limiting a method to one dataset would be omitting a large amount of data. However, thus far, it is impossible to use them all at once, as there is no unique fine label set. 

\begin{table}[h]
\scriptsize
    \centering
    \begin{tabular}{|c|c|c|c|}
        \hline Dataset & Size & Scene type & \#Labels \\\hline
        KITTI-360 \cite{360} & 80K & Suburbs & 27\\\hline
        nuScenes \cite{nuscenes2019} & 35K &  Urban & 16\\\hline
        SemanticKITTI \cite{behley2019iccv} & 23K &  Suburbs & 19\\ \hline
        PandaSet \cite{panda} & 6K &  Urban \& Road & 37\\ \hline
        SemanticPOSS \cite{poss} & 3K &  Campus & 13\\ \hline
    \end{tabular}
    \caption{Summary of the datasets used in the different experiments. The size corresponds to the number of labeled scans. Labels correspond to the labels used at the inference time.}
    \label{tab:dataset}
\end{table}

Despite this obvious obstacle, a study of the datasets shows that labels are organized in a tree-like structure with more or less refined labels, similar to the one highlighted by \cite{8659045}. Moreover, higher levels of trees are more uniform across datasets.

\subsection{Coarse Labels}
\label{sec:coarse_lab}
Based on these observations, we propose a set of \textbf{coarse labels} such that we can easily map \textit{fine} labels from any dataset to this one. These coarse labels define categories that retain information relevant to the end task, are consensual across all existing and future such datasets, and display little ambiguity; furthermore, they come at no additional annotation cost. 

Besides these technical advantages, we believe that those coarse labels can still represent effectively the major elements of a typical road scene such that a model trained with them would still be relevant for autonomous driving. The coarse labels we propose are as follows:
\begin{itemize}
    \item Driveable ground covers areas where cars are expected to be driving.
    \item Other ground corresponds to other large planar surfaces, where pedestrians can be found.
    \item Structure corresponds to fixed human-made structures, which can be hard to subdivide into smaller groups.
    \item Vehicles are the typical road users.
    \item Nature corresponds to the vegetation-like components of the scene.
    \item Living being groups humans and animals alike, who can interact with all the other elements.
    \item Dynamic objects are objects that can typically move, such as strollers.
    \item Static objects are objects that are fixed, such as poles.
\end{itemize}

An illustration of the labels and the mapping of the fine labels from SemanticKITTI to the proposed label set can be found in \autoref{fig:coarse_label}. All mappings are available in our public GitHub repository\footnote{\url{https://github.com/JulesSanchez/CoarseLabel}}.

\subsection{COLA (Coarse Label pre-training)}
Given this new coarse label set, we can perform semantic segmentation on all the data at once. This supervised learning is the proposed COarse LAbel (COLA) pre-training presented in the introduction. To reuse such a pre-trained model, only the last layer of the model needs to be replaced for finetuning, to account for the new number of labels (see \autoref{fig:pipe}). Due to the size of the architecture, no layers need to be frozen at the time of finetuning. In this way, COLA is expected to learn representations tailored for the specific task of autonomous driving semantic segmentation. The validation mIoU of the pre-training reaches between $65$\% and $78$\%, depending on the target dataset, which is sufficient to conclude that the architecture has learned relevant features, confirmed in \autoref{sec:ptfa}.

In practice, COLA took the equivalent of 10 GPU days of training time, resulting in a much shorter pre-training procedure than SPC.

We expect COLA to perform well for several reasons. First, seeing various sensor and environment settings in its training set will help generalize to new scenes and sensors. Second, aligning the pre-training and finetuning tasks to be very similar ensures the relevance of the learned features, both geometrically and semantically.

\subsection{Pre-training feature analysis}
\label{sec:ptfa}
In this section, we proceed to a feature analysis of the last layer, before the classification of a SRU-Net model after the COLA pre-training. We base the analysis on the SemanticPOSS dataset to study COLA when the largest amount of pre-training data is available. 

\begin{figure}[h]
        \centering
        \subfloat[t-SNE analysis of COLA \\ colored with coarse labels]{
            \includegraphics[width=.4\linewidth]{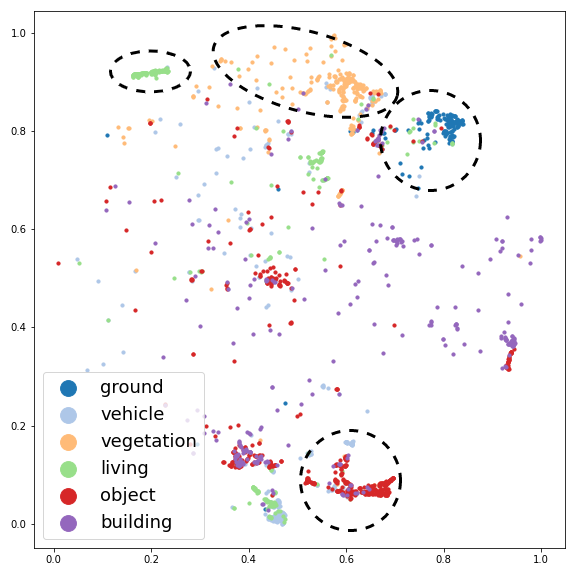}
        }
        \subfloat[t-SNE analysis of COLA \\ colored with fine labels]{
            \includegraphics[width=.4\linewidth]{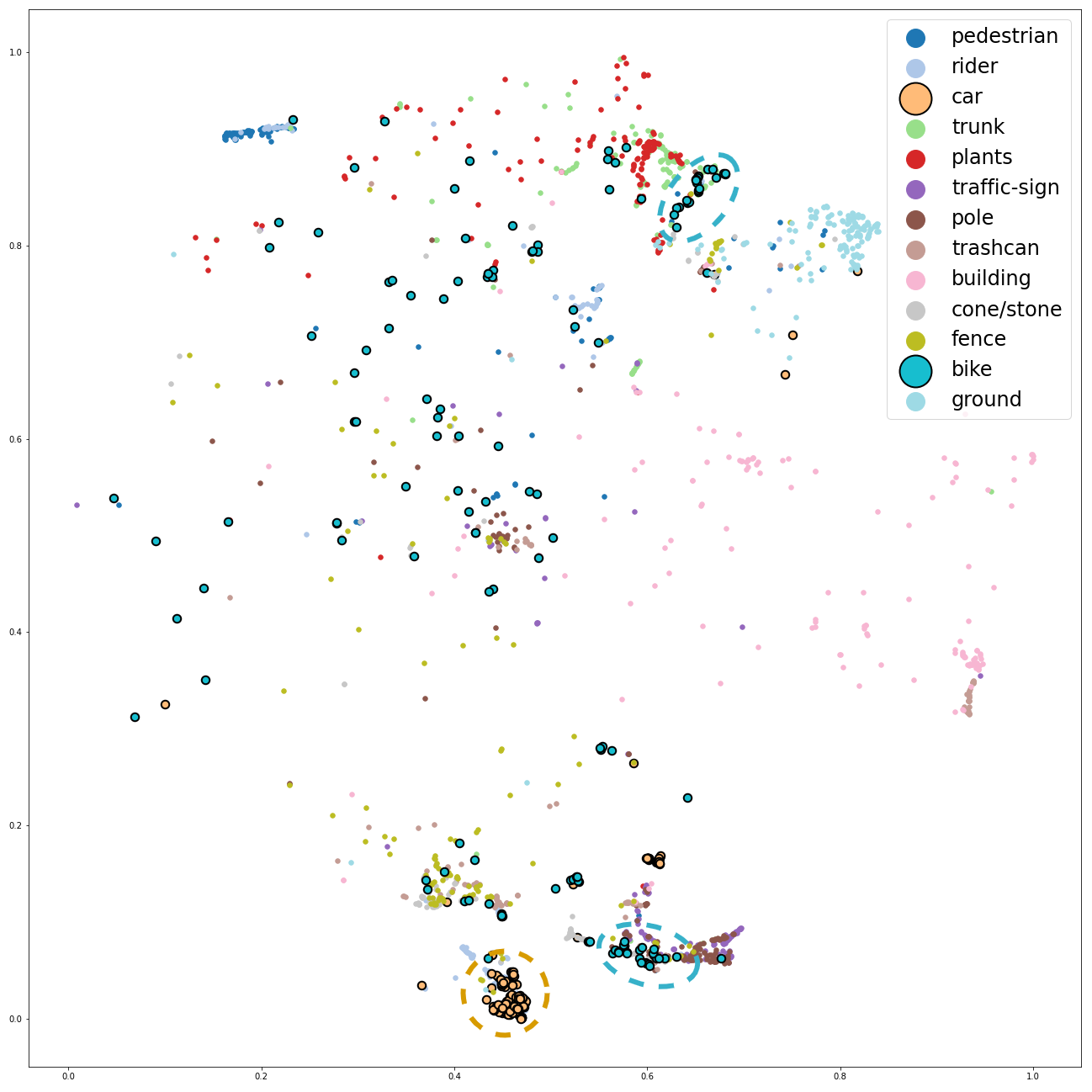}
        } 
    \caption{t-SNE analysis of the last feature layers of SRU-Net trained with COLA. Inference performed on SemanticPOSS.}
    \label{fig:tsne_pretraining}
\end{figure}

To conduct the analysis, we give point clouds extracted from SemanticPOSS until we reach at least 200 points for each ground truth category and extract their associated last layer features. We then do a t-SNE analysis and plot the 2D projection in \autoref{fig:tsne_pretraining}, where points are colored according to either their coarse ground truth or fine ground truth. We observe the ability of the model to correctly cluster points in both the coarse label set and in the fine label set, even though it was not yet finetuned and had never seen any SemanticPOSS data. This demonstrates the ability of COLA to generalize the segmentation to unseen data and furthermore shows the relevance of learning coarse labels even when fine labels are the end goal.

We focus on a particular case of the COLA t-SNE analysis in \autoref{fig:tsne_pretraining}. In the right image, we highlighted (points are circled in black) points belonging to either car or bike categories. We observe a highly desirable pattern, which is that COLA learned to distinguish bikes from cars feature-wise, even if they were not annotated as such in the coarse labels. This highlights that the pre-training method leverages both semantics and geometry, and thus, COLA does not rely on a perfect intermediate label set to learn relevant features.

\section{Experiments}
\label{sec:expe}
In this section, we describe the experimental setup used to evaluate the performance of the proposed coarse label pre-training (COLA). 
\subsection{Experimental cases}

For all our experiments, we split the data into two categories: the \textbf{target} data, which is the data we used for finetuning, on which we compute performance metrics, and the \textbf{pre-training} data. KITTI-360 is not used for pre-training in the case of the SemanticKITTI target as there is a significant geographic overlap of their scenes.

The four cases we chose correspond to experimenting on the two reference datasets in the community (nuScenes and SemanticKITTI), as well as on two limited data availability setups (SemanticPOSS and partial nuScenes). Additional details can be found in \autoref{tab:data_dist}.
Partial nuScenes is a homemade set of datasets, which corresponds to subsets of nuScenes to provide access to a wider range of dataset sizes; otherwise, we would be limited to SemanticPOSS.

As highlighted in \autoref{tab:dataset}, target datasets present different scene types, and furthermore, nuScenes and SemanticKITTI are acquired with sensors of very different resolutions. In this way, we can evaluate the methods in many settings.

We experiment on two architectures, SPVCNN \cite{spvnas} and SRU-Net \cite{mink} which are staples of the autonomous driving semantic segmentation field, either as state-of-the-art architecture or typical backbone.

To study COLA, we compute the mean Intersection over Union (mIoU) computed on an extracted test set. All experiments are performed on NVIDIA GeForce RTX 3090 GPUs.

\begin{table}[h!]
\scriptsize
    \centering
    \begin{tabular}{|c|c|c|c|}
    \hline
        Pre-training data &\begin{tabular}{@{}c@{}} \# Scans for\\ pre-training\end{tabular}  & Target &  \begin{tabular}{@{}c@{}} \# Size ratio\end{tabular} \\
        \hline
         \begin{tabular}{@{}c@{}}SK + K360 + SP + PS\end{tabular} & 110K & NS & 25\%\\\hline
         \begin{tabular}{@{}c@{}}SK + K360 + PS + NS\end{tabular} & 140K & SP & 2\%\\
        \hline
         \begin{tabular}{@{}c@{}}SP + NS + PS \end{tabular}& 45K & SK & 33\%\\
         \hline 
         \begin{tabular}{@{}c@{}}SK + K360 + SP + PS\end{tabular}& 110K & \specialcell{PartialNS}  & \specialcell{2.5\%-25\%}\\
         \hline 
         \begin{tabular}{@{}c@{}}SK + K360 + SP + PS\end{tabular}& 110K & \specialcell{PartialSP}  & \specialcell{0.7\%}\\
         \hline
    \end{tabular}
    \caption{Datasets used for pre-training depending on the target and the size ratio between the target set and the pre-training set. The various datasets are SemanticKITTI (SK), KITTI-360 (K360), nuScenes (NS), SemanticPOSS (SP) and PandaSet (PS)}
    \label{tab:data_dist}
\end{table}

\subsection{Partial nuScenes and partial SemanticPOSS}

We introduced partial nuScenes previously. Partial nuScenes is a subset of nuScenes, based on the number of scenes. We propose three levels: 10\%, 25\%, and 50\%, which represent 95, 235, and 470 scenes (respectfully 1900, 4700, and 9400 scans), respectively, for training and validation, which are split in the same proportion as the full nuScenes 70/30.

We similarly introduce partial SemanticPOSS, composed of one sequence (00) for training and one for validation.

The objective of introducing such datasets is the ability to study the effect of COLA for a larger variety of scans available during the finetuning stage. We selected SemanticPOSS and nuScenes as they are the setup in which the pre-training data available is the largest. 

\subsection{Networks and parameters}

We experimented on two architectures (SPVCNN \cite{spvnas} and SRU-Net \cite{mink}). All parameters were chosen based on the from-scratch training and simply re-used in the pre-trained setup.

\subsubsection{SRU-Net}

For the COLA pre-training, we used the SGD optimizer, with an initial learning rate of 0.4, a momentum of 0.9, and a cosine annealing scheduler. We had a batch size of 48 over 10 epochs. 

For finetuning, whether the model has been pre-trained or not, we use the same parameters. We used the SGD optimizer, with an initial learning rate of 0.8, a momentum of 0.9, and a cosine annealing scheduler. We had a batch size of 36 over 30 epochs.

The model used is the same as the one used in \cite{csc2021}.

\subsubsection{SPVCNN}

For the COLA pre-training and the finetuning, we used the SGD optimizer, with an initial learning rate of 0.24, a momentum of 0.9, and a cosine warmup scheduler. We had a batch size of 16 over 15 epochs.

The model used is the one provided by the original authors in the official GitHub repository \cite{spvnas}.

All models are trained with the mixed Lovasz loss function and following the same data augmentation as \cite{mink}.

\section{Results}
\label{sec:res}

In this section, we show the results from the thorough experiments described in the previous sections. They are divided into two parts. First, we look at results on full datasets (SemanticKITTI, SemanticPOSS, and nuScenes), and on each network architectures, and then, we investigate more precisely the case of limited data with partial nuScenes. In each case, we show the mIoU over a test set, which we extract from the training set, representing between 10\% and 20\% of the available data. We compare all COLA results to the no pre-training case. 

All mIoU gains that fall within  $\pm0.5$ are due to the randomness of the training rather than the effect of a method.

\subsection{Results on SemanticKITTI, nuScenes, and SemanticPOSS}

\begin{table}[h]
\scriptsize
    \centering
    \begin{tabular}{|c|c|c||c|c|}
    \hline
      &\multicolumn{2}{c||}{SRU-Net}&\multicolumn{2}{c|}{SPVCNN} \\ \hline
         &  \specialcell{No pre-\\training} & \specialcell{COLA\\(Ours)}&  \specialcell{No pre-\\training} & \specialcell{COLA\\(Ours)}\\ \hline
        nuScenes& 67.2 & \textbf{69.3 (+2.1)} & 65.9 & \textbf{66.2 (+0.3)}\\ \hline
        SemanticKITTI& 50.5 & \textbf{52.4 (+1.9)} & 57.7 & \textbf{58.4 (+0.7)}\\ \hline
        SemanticPOSS& 55 & \textbf{55.8 (+0.8)} & 49.9 & \textbf{53.8 (+3.9)}\\ \hline
    \end{tabular}
    \caption{Test set mIoU on each target dataset. No pre-training, and COLA pre-training.}
    \label{tab:tab_SRU-Net}
\end{table}

Overall, the models pre-trained with the proposed COLA pre-training show significant improvement compared with the other methods, with a gain of up to +$2\%$ for SRU-Net and up to +$4\%$ for SPVCNN \autoref{tab:tab_SRU-Net}, in every case displaying significant improvement when compared with naive contrastive approaches (\autoref{tab:tab_SRU-Net_contr}).

\begin{figure}[h]
    \centering
    \includegraphics[width=0.9\linewidth]{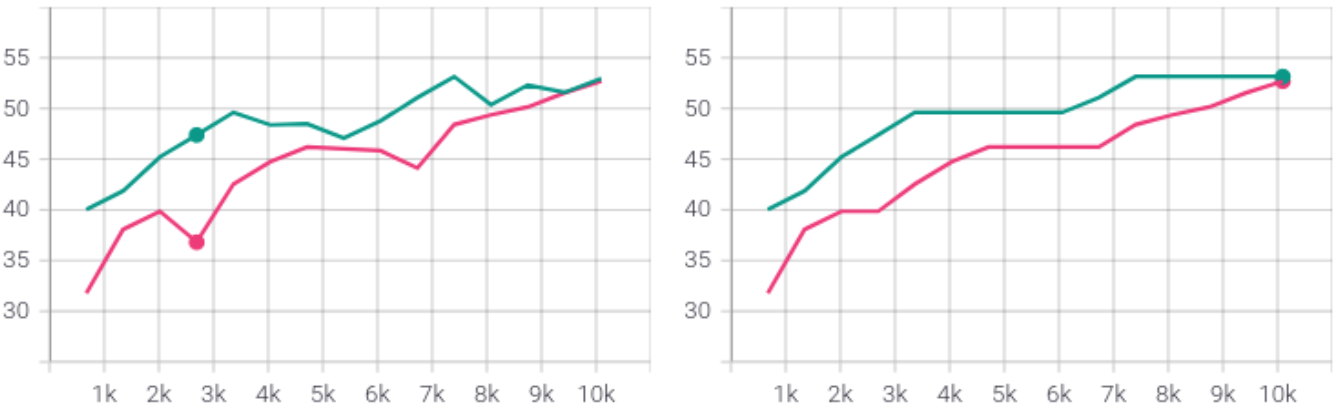}
    \caption{Validation mIoU curve for SRU-Net finetuned (green) and trained from scratch (pink) on SemanticPOSS. On the left, mIoU at each iteration, on the right best mIoU so far.}
    \label{fig:convergence speed}
\end{figure}

In \autoref{fig:convergence speed}, we study an example were the finetuned model show only a small improvement over the training from scratch. We see that, while the final mIoU are similar, it is reached much faster when pre-trained, and early epochs show some significant improvements.

\subsection{Results on partial nuScenes and partial SemanticPOSS}
\begin{table}[h]

\scriptsize

\scalebox{0.95}{
\begin{tabular}{|c|c||c|c||c|c|}
    \hline
    \% scenes& \# of scans &\multicolumn{2}{c||}{SRU-Net}&\multicolumn{2}{c|}{SPVCNN}\\\hline
    & & \specialcell{No\\pre-training} & \specialcell{COLA\\(Ours)} & \specialcell{No\\pre-training} & \specialcell{COLA\\(Ours)}\\\hline
        10\%& 1900 & 46.2 & \textbf{48.4 (+2.2)} & 44.9 & \textbf{51.3 (+6.4)} \\\hline
        25\%& 4700 & 58.0 & \textbf{60.0 (+2.0)} &52.5 & \textbf{55.6 (+3.1)}\\\hline
        50\%& 9400 & 62.4 & \textbf{65.0 (+2.6)} &56.8 & \textbf{57.2 (+0.4)}\\\hline
        100\%& 16000 & 67.2 & \textbf{69.3 (+2.1)} &65.9 & \textbf{66.2 (+0.3)}\\\hline
\end{tabular}}

\caption{Test set mIoU on each partial nuScenes. No pre-training, and COLA pre-training.}
\label{tab:NS_limited}
\end{table}

\begin{table}[h]

\scriptsize
\scalebox{0.95}{
\begin{tabular}{|c||c|c||c|c|}
    \hline
    \# of scans &\multicolumn{2}{c||}{SRU-Net}&\multicolumn{2}{c|}{SPVCNN}\\\hline
    & \specialcell{No pre-training} & \specialcell{COLA (Ours)} & \specialcell{No pre-training} & \specialcell{COLA (Ours)}\\\hline
    500 & 35.2 & \textbf{37.9 (+2.7)} & 32.9 & \textbf{36.8 (+3.9)} \\\hline
    1500 (full) & 55 & \textbf{55.8 (+0.8)} & 49.9 & \textbf{53.8 (+3.9)} \\\hline

\end{tabular}}

\caption{Test set mIoU on partial SemanticPOSS. No pre-training and COLA pre-training.}
\label{tab:SP_limited}
\end{table}

We compiled results for partial nuScenes for both architectures in \autoref{tab:NS_limited} and the ones for partial SemanticPOSS in \autoref{tab:SP_limited}. We confirmed the previous conclusions that the proposed COLA pre-training improves the performance significantly, whatever the architecture, when a small dataset is available. This is even more significant for SPVCNN.

\subsection{Qualitative results}

\begin{figure}[h]
        \centering
        \subfloat[Results for \\ no pre-training]{
            \includegraphics[width=.4\linewidth]{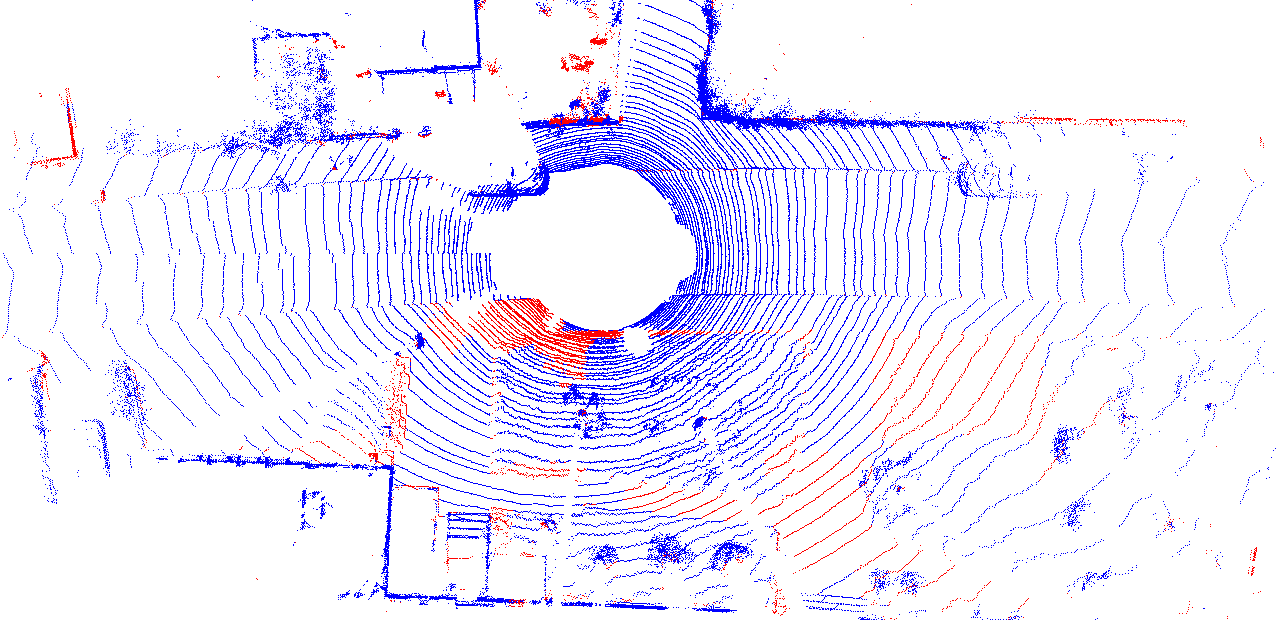}
        }
        \subfloat[Results for \\ COLA training]{
            \includegraphics[width=.4\linewidth]{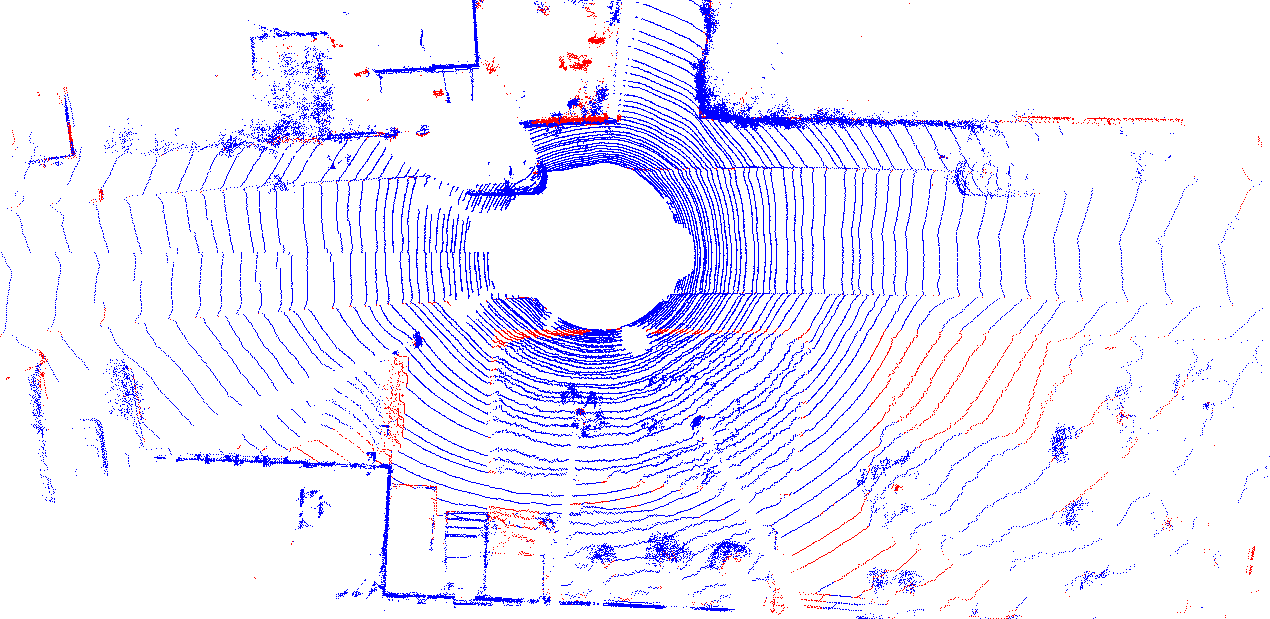}
        } 
    \caption{Example of results on SemanticKITTI after finetuning with SRU-Net. In blue, points with correct semantic segmentation. In red, errors.}
    \label{fig:semantic_quali}
\end{figure}

We show some qualitative results on SemanticKITTI with SRU-Net in \autoref{fig:semantic_quali}. Segmentation errors are red, and we can see a significant decrease in mistakes thanks to COLA pre-training especially for buildings and vegetation. Zoom in for a better visualization. 

\subsection{Ablation study}

In the ablation study presented in \autoref{tab:ab_study}, we analyze the importance of the different steps that are part of our proposed COLA pre-training. We conduct our analysis with SemanticPOSS as the target dataset as this is the one with largest size ratio (see \autoref{tab:data_dist}). We are comparing four cases: no pre-training, pre-training with only KITTI-360 labeled with its original fine labels, pre-training with only KITTI-360 labeled with our proposed coarse labels, and finally, applying the full COLA, that is, all the pre-training data labeled with the coarse labels. 

First, we arrive at the same conclusions as the one presented in PointContrast \cite{xie2020pointcontrast}. In the case of 3D scene understanding, pre-trainings still provide a substantial gain compared with training from scratch, and the data ceiling is not reached yet, as shown by the increase in performance between COLA and pre-training with only KITTI-360 (+1.4 mIoU) (\autoref{tab:ab_study}).

The second highlight is the impact of the coarse labels compared with the fine labels. We demonstrate an improvement in using the coarse labels rather than the fine labels as a pre-training task (+0.6 mIoU). This means than beside providing a simpler mapping in between label sets, it also serves a purpose model-wise. This stems from the simplicity of the labels provided by the coarse labels. When finetuning the model does not need to unlearn anything, unlike the pre-training with the fine labels, where some labels and objects do not exist anymore.

Finally, we compare our approach (COLA) with a more naive pre-training approach that would have a classification module for each of the pre-training datasets rather than having a unified classification module. In this way, each dataset is trained with its fine labels rather than using the coarse label set. This method is referred to MH (multi-head) in \autoref{tab:ab_study}. COLA demonstrates a significant improvement (+1.1 mIoU) over MH and highlights the effectiveness of using an intermediate label set rather than naively using fine label sets.

In \autoref{sec:coarse_lab}, we proposed a coarse label set we deemed representative of autonomous driving data. We showed that it managed to learn useful geometry information beyond the annotated coarse label: nevertheless, the choice of the label set could still be discussed. In \autoref{tab:ab_study_n_labels}, we study the impact of the coarse label set choice: more precisely, we compare the results when we use coarser category (5 instead of 8 by grouping driveable ground and other ground, and grouping dynamic and static objects and structure) or finer categories (10 categories instead of 8 by subdividing static objects into poles and other objects, and subdividing vehicles into 2-wheeled and 4-wheeled vehicles). We can see that the pre-training is robust to the number and choice of coarse labels, allowing in all cases a gain of about +2\% mIoU, with a slight advantage for our 8 categories.

\begin{table}[h]
\scriptsize
    \centering
    \begin{tabular}{|c||c|}
    \hline
        Methodology & mIoU \\ \hline
        Only finetuned on SemanticPOSS  = (1)& 49.9 \\ \hline
        Pre-trained with KITTI-360 (fine labels) + (1)  & 51.8 (+1.9) \\ \hline
        Pre-trained with KITTI-360 (coarse labels) + (1)  & 52.4 (+2.5)\\\hline
        Pre-trained with MH (all datasets) + (1)  & 52.7 (+2.8) \\\hline
        Pre-trained with COLA (all datasets) + (1)  & 53.8 (+3.9)\\ \hline
    \end{tabular}
    \caption{Ablation study of the  method focusing on the impact of the coarse labels and the amount of pre-training data with SPVCNN. The target dataset on which the test set mIoU is computed is SemanticPOSS.}
    \label{tab:ab_study}
\end{table}

\begin{table}[h]
\scriptsize
    \centering
    \begin{tabular}{|c||c|}
    \hline
        Methodology & mIoU \\ \hline
        Only finetuned on SemanticPOSS = (1) & 49.9 \\ \hline
        Pre-trained with KITTI-360 with 5 coarse labels + (1)  & 52.2 (+2.3) \\\hline
        Pre-trained with KITTI-360 with 10 coarse labels + (1)  & 52.1 (+2.2) \\\hline
        Pre-trained with KITTI-360 with the same labels as COLA + (1)  & 52.4 (+2.5) \\\hline
    \end{tabular}
    \caption{Ablation study of the  method focusing on the impact of the number of coarse labels with SPVCNN. The target dataset on which the test set mIoU is computed is SemanticPOSS.}
    \label{tab:ab_study_n_labels}
\end{table}

\section{Conclusion}

In this paper, we proposed a new supervised pre-training method for autonomous driving semantic segmentation in order to leverage existing annotations and extract meaningfully semantic and geometric information from the available data.

We demonstrated that the proposed COLA method shows great promises. Specifically, it works very well when the amount of target data is limited, but the proposed method is not restricted to this use case. Furthermore, we have shown the proposed method works on various data setups and manages to improve performance even when the target setup is not seen during the pre-training. COLA needs fewer raw data than typical contrastive approaches followed thus far, in addition to needing significantly less computational power and time.

\bibliographystyle{unsrt}
\bibliography{bibliography}

\end{document}